\documentclass[11pt,letterpaper]{article}
\usepackage{emnlp2017}
\usepackage{times}
\usepackage{latexsym}
\usepackage{amsmath}
\usepackage{graphicx}
\usepackage{multirow}
\usepackage{url}
\usepackage[T1]{fontenc}

\emnlpfinalcopy

\def\emnlppaperid{***}

\title{SYSTRAN Purely Neural MT Engines for WMT2017}

\author{ 
{\bf Yongchao Deng},
{\bf Jungi Kim}, 
{\bf Guillaume Klein}, 
{\bf Catherine Kobus}, 
{\bf Natalia Segal}, \\
{\bf Christophe Servan}, 
{\bf Bo Wang},  
{\bf Dakun Zhang},
{\bf Josep Crego},
{\bf Jean Senellart} \\ \\
  {\tt firstname.lastname@systrangroup.com} \\
    SYSTRAN / 5 rue Feydeau, 75002 Paris, France 
}

\date{}

\begin{document}

\maketitle

\begin{abstract}

This paper describes SYSTRAN's systems submitted to the WMT 2017 shared news translation task for English-German, in both translation directions. Our systems are built using OpenNMT\footnote{\url{http://opennmt.net}}, an open-source neural machine translation system, implementing sequence-to-sequence models with LSTM encoder/decoders and attention. We experimented using monolingual data automatically back-translated. Our resulting models are further hyper-specialised with an adaptation technique that finely tunes models according to the evaluation test sentences.

\end{abstract}

\section{Introduction}
\label{sec:intro}

We participated in the WMT 2017 shared news translation task on two different translation directions: English$\rightarrow$German and German$\rightarrow$English. 

The paper is structured as follows: Section \ref{sec:neural} overviews our neural MT engine. Section \ref{sec:exp} describes the set of experiments carried out to build the English$\rightarrow$German and German$\rightarrow$English neural translation models. Experiments and results are detailed in Section \ref{sec:exp}. Finally, conclusions are drawn in Section \ref{sec:conclusions}.

\section{Neural MT System}
\label{sec:neural}

Neural machine translation (NMT) is a new methodology for machine translation that has led to remarkable improvements, particularly in terms of human evaluation, compared to rule-based and statistical machine translation (SMT) systems \cite{DBLP:journals/corr/CregoKKRYSABCDE16,GNMT}. NMT has now become a widely-applied technique for machine translation, as well as an effective approach for other related NLP tasks such as dialogue, parsing, and summarisation.

\begin{figure*}[h]
   \includegraphics[width=0.99\textwidth]{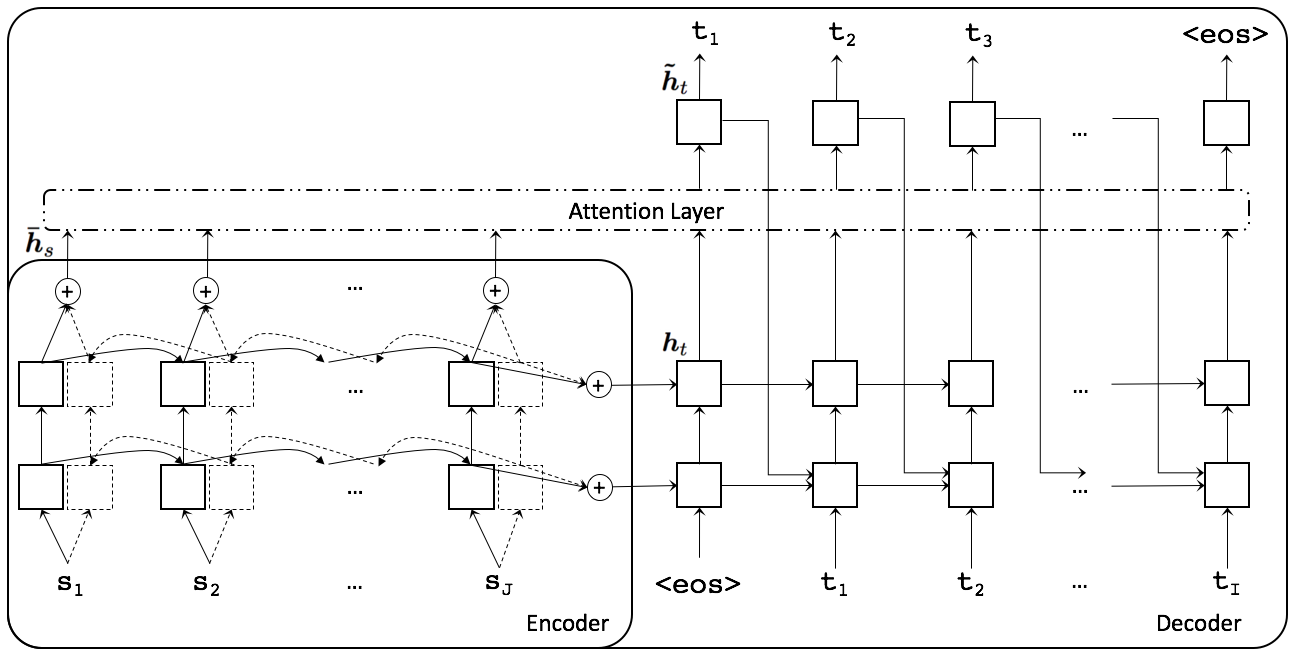}
\caption{Schematic view of our MT network.}
\label{opennmt}
\end{figure*}

Our NMT system \cite{KleinKDSR17} follows the architecture presented in ~\cite{DBLP:journals/corr/BahdanauCB14}. It is implemented as an encoder-decoder network with multiple layers of a RNN with Long Short-Term Memory (LSTM) hidden units ~\cite{DBLP:journals/corr/ZarembaSV14}. Figure \ref{opennmt} illustrates an schematic view of the MT network. 

Source words are first mapped to word vectors and then fed into a bidirectional recurrent neural network (RNN) that reads an input sequence $s = (s_1,...,s_J)$. Upon seeing the \texttt{<eos>} symbol, the final time step initialises a target RNN. The decoder is a RNN that predicts a target sequence $t = (t_1, ..., t_I)$, being $J$ and $I$ respectively the source and target sentence lengths. Translation is finished when the decoder predicts the \texttt{<eos>} symbol.

The left-hand side of the figure illustrates the bidirectional encoder, which actually consists of two independent LSTM encoders: one encoding the normal sequence (solid lines) that calculates a forward sequence of hidden states 
$(\overrightarrow{h_1}, ..., \overrightarrow{h_J})$, the second encoder reads the input sequence in reversed order (dotted lines) and calculates the backward sequence $(\overleftarrow{h_1},..., \overleftarrow{h_J})$. 
The final encoder outputs $(\overline{h}_1, ..., \overline{h}_J)$ consist of the sum of both encoders final outputs. 
The right-hand side of the figure illustrates the RNN decoder. Each word $t_i$ is predicted based on a recurrent hidden state $h_i$ and a context vector $c_i$ that aims at capturing relevant source-side information.

\begin{figure}[h]
   \includegraphics[width=0.48\textwidth]{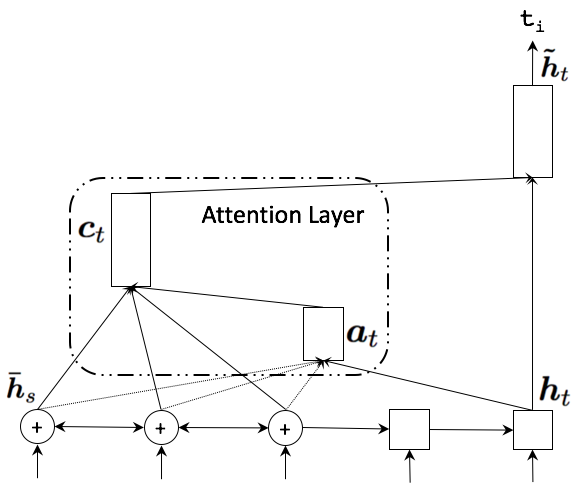}
\caption{Attention layer of the MT network.}
\label{attention}
\end{figure}

Figure \ref{attention} illustrates the attention layer. It implements the "general" attentional architecture from ~\cite{luong-pham-manning:2015:EMNLP}. 
The idea of a global attentional model is to consider all the hidden states of the encoder when deriving the context vector $c_t$.
Hence, global alignment weights $a_{t}$ are derived by comparing the current target hidden state $h_t$ with each source hidden state $\overline{h}_s$:

\begin{equation*}
a_{t}(s) = \frac{exp(score(h_t,\overline{h}_s))}{\sum_{s'} exp(score(h_t,\overline{h}_{s'}))}
\end{equation*}

with the content-based score function:
\begin{equation*}
score(h_t,\overline{h}_s) = h_t^T W_a \overline{h}_s
\end{equation*}

Given the alignment vector as weights, the context vector $c_t$ is computed as the weighted average over all the source hidden states.

Note that for the sake of simplicity figure~\ref{opennmt} illustrates a two-layers LSTM encoder/decoder while any arbitrary number of LSTM layers can be stacked. 
More details about our system can be found in \cite{DBLP:journals/corr/CregoKKRYSABCDE16}. 

\section{Experiments}
\label{sec:exp}

In this section we detail the corpora and training experiments used to build our English$\leftrightarrow$German neural translation models.

\subsection{Corpora}
\label{ssec:corpora}

We used the parallel corpora made available for the shared task: {\it Europarl v7}, {\it Common Crawl corpus}, {\it News Commentary v12} and {\it Rapid corpus of EU press releases}. Both English and German texts were preprocessed with standard tokenisation tools. German words were further preprocessed to split compounds, following a similar algorithm as the built-in for Moses. Additional monolingual data was also used for both German and English available for the shared task: {\it News Crawl: articles from 2016}. Basic statistics of the tokenised data are available in Table \ref{tab:data}.

\begin{table} [h]
\centerline{
\scalebox{1}{
\begin{tabular}{|c|c|c|c|c|}
\hline
 & \#sents & \#words & vocab. & $L_{mean}$ \\
\hline  
\multicolumn{5}{l}{{\it Parallel}} \\
\hline
En & 4.6M & 103.7M & 627k & 22.6\\
De & 4.6M & 104.5M & 836k & 22.8\\
\hline  
\multicolumn{5}{l}{{\it Monolingual}} \\
\hline  
En & 20,6M & 463,6M & 1.18M & 22.5\\
\hline
De & 34,7M & 620,8M & 3.36M & 17.8\\
\hline 
\end{tabular}}}
\caption{\label{tab:data} {\it English-German parallel and monolingual corpus statistics. $L_{mean}$ indicates mean sentence lengths. M stand for millions, k for thousands.}}
\end{table}

We used a byte pair encoding technique\footnote{\url{https://github.com/rsennrich/subword-nmt}} (BPE) to segment word forms and achieve open-vocabulary translation with a fixed vocabulary of $30,000$ source and target tokens. 
BPE was originally devised as a compression algorithm, adapted to word segmentation~\cite{Sennrich2016}. 
It recursively replaces frequent consecutive bytes with a symbol that does not occur elsewhere. 
Each such replacement is called a merge, and the number of merges is a tuneable parameter. 
Encodings were computed over the union of both German and English training corpora after preprocessing, aiming at improving consistency between source and target segmentations.

Finally, case information was considered by the network as an additional feature. It allowed us to work with a lowercased vocabulary and treat re-casing as a separate problem~\cite{DBLP:journals/corr/CregoKKRYSABCDE16}.

\subsection{Training Details}
\label{ssec:training}

All experiments employ the NMT system detailed in Section \ref{sec:neural}. The encoder and the decoder consist of a four-layer stacked LSTM with $1,000$ cells each. We use a bidirectional RNN encoder.
Size of word embedding is $500$ cells. We use stochastic gradient descent, a minibatch size of $64$ sentences and $0.3$ for dropout probability. Maximum sentence length is set to $80$ tokens. All experiments are performed on NVidia GeForce GTX 1080 on a single GPU per optimisation work. Newstest2008 (2008) is employed as validation test set and newstest from 2009 to 2016 (2009-16) as internal test sets. 

\subsubsection{Training on parallel data}

Table \ref{tab:results1} outlines training work. All parallel data ({\bf P}) is used on each training epoch. Row LR indicates the learning rate value used for each epoch. Note that learning rate was initially set to $1.0$ during several epochs until no or little perplexity (PPL) reduction is measured on the validation set. Afterwards, additional epochs are performed with learning rate decayed by $0.7$ at each epoch. BLEU score (averaged over the eight internal test sets) after each training epoch is also shown. Note that all BLEU scores shown in this paper are computed using \texttt{multi-bleu.perl}\footnote{\url{https://github.com/moses-smt/mosesdecoder/blob/master/scripts/generic/multi-bleu.perl}}. Training time per epoch is also shown in row Time measured in number of hours.

\begin{table*}[h]
\begin{center}
\scalebox{.8}{
\begin{tabular}{|lccccccccccccc|}
  \hline
  Epoch & 1 & 2 & 3 & 4 & 5 & 6 & 7 & 8 & 9 & 10 & 11 & 12 & 13 \\
  \hline
  \multicolumn{14}{l}{German$\rightarrow$English} \\
  \hline
  Data   & P & P & P & P & P & P & P & P & \multicolumn{1}{c|}{P} & P & P & P & P \\
  Time (hours)   & 24 & 24 & 24 & 24 & 24 & 24 & 24 & 24 & \multicolumn{1}{c|}{24} & 24 & 24 & 24 & 24 \\ 
  LR      & $1.0$ & $1.0$ & $1.0$ & $1.0$ & $1.0$ & $1.0$ & $1.0$ & $1.0$ & \multicolumn{1}{c|}{$1.0$} & $0.7^1$ & $0.7^2$ & $0.7^3$ & $0.7^4$  \\
  PPL (2008)  & $20.90$ & $17.01$ & $15.38$ & $14.67$ & $14.18$ & $13.75$ & $13.57$ & $13.29$ & \multicolumn{1}{c|}{$13.00$} & $12.47$ & $12.05$ & $11.49$ & $11.40$ \\
  BLEU (2009-16) & $20.07$ & $22.06$ & $23.02$ & $24.17$ & $24.59$ & $24.40$ & $24.99$ & $25.11$ & \multicolumn{1}{c|}{$25.42$} & $25.65$ & $26.14$ & $26.48$ & $26.87$ \\
  \hline
  \multicolumn{14}{l}{English$\rightarrow$German} \\
  \hline
  Data             & P & P & P & P & P & P & P & \multicolumn{1}{c|}{P} & P & P & P & P & P \\
  Time (hours)   & 24 & 24 & 24 & 24 & 24 & 24 & 24 & \multicolumn{1}{c|}{24} & 24 & 24 & 24 & 24 & 24 \\ 
  LR                & $1.0$ & $1.0$ & $1.0$ & $1.0$ & $1.0$ & $1.0$ & $1.0$ & \multicolumn{1}{c|}{$1.0$} & $0.7^1$ & $0.7^2$ & $0.7^3$ & $0.7^4$ & $0.7^5$ \\
  PPL (2008)  & $20.85$ & $16.52$ & $14.84$ & $13.89$ & $13.62$ & $13.13$ & $12.59$ & \multicolumn{1}{c|}{$12.66$} & $11.72$ & $11.20$ & $10.94$ & $10.75$ & $10.55$ \\
  BLEU (2009-16) & $15.63$ & $17.41$ & $18.85$ & $19.61$ & $19.92$ & $20.38$ & $20.34$ & \multicolumn{1}{c|}{$20.55$} & $21.13$ & $21.63$ & $21.70$ & $22.22$ & $22.50$ \\
  \hline
\end{tabular}
}
\end{center}
\caption{\label{tab:results1} {\it Training on parallel data.}}
\end{table*}

As expected, a perplexity reduction is observed for the initial epochs, until epochs $9$ (German$\rightarrow$English) and $8$ (English$\rightarrow$German) where little or no improvement is observed. The decay mode is then started allowing to further boost accuracy (between $1.5$ and $2.0$ BLEU points) after $5$ additional epochs.

\subsubsection{Training on parallel and synthetic data}

Following \cite{sennrichMono}, we selected a subset of the available target-side in-domain monolingual corpora, translate it into the source side (back-translate) of the respective language pair, and then use this synthetic parallel data for training. The best performing models for each translation direction (epoch 13 on Table \ref{tab:results1} of both translation directions) were used to back-translate monolingual data. \cite{sennrichMono} motivate the use of monolingual data with domain adaptation, reducing overfitting, and better modelling of fluency.

Synthetic corpus was then divided into $i$ different splits containing each $4.5$ million sentence pairs (except for the last split that contains less sentences). Table \ref{tab:results2} shows continuation of the training work using at each epoch the union of the entire parallel data together with a split of the monolingual back-translated data ({\bf P+M$_i$}). Hence, balancing the amount of reference and synthetic data, summing up to around $9$ million sentence pairs per epoch. Note that training work described in Table \ref{tab:results2} is built as continuation of the model at epoch $13$ on Table \ref{tab:results1}. 
Table \ref{tab:results2} shows also BLEU scores over newstest2017 for the best performing network. 

\begin{table*}[h]
\begin{center}
\scalebox{.74}{
\begin{tabular}{|lccccccccccccc}
  \hline
  Epoch & 1 & 2 & 3 & 4 & 5 & 6 & 7 & 8 & 9 & 10 & 11 & 12 & \multicolumn{1}{c|}{13} \\
  \hline
  \multicolumn{14}{l}{German$\rightarrow$English} \\
  \cline{1-11}
  Data   & P+M$_1$ & P+M$_2$ & P+M$_3$ & P+M$_4$ & \multicolumn{1}{c|}{P+M$_5$} & P'+M' & P'+M' & P'+M' & P'+M' & \multicolumn{1}{c|}{P'+M'} & & & \\
  Time (hours)   & $45$ & $45$ & $45$ & $45$ & \multicolumn{1}{c|}{$32$} & $25$ & $25$ & $25$ & $25$ & \multicolumn{1}{c|}{$25$} \\ 
  LR      & $1.0$ & $1.0$ & $1.0$ & $1.0$ & \multicolumn{1}{c|}{$1.0$} & $0.7^1$ & $0.7^2$ & $0.7^3$ & $0.7^4$ & \multicolumn{1}{c|}{$0.7^5$} & & & \\
  PPL (2008)  & $13.33$ & $13.23$ & $13.26$ & $1347$ & \multicolumn{1}{c|}{$12.63$} & $12.25$ & $11.87$ & $11.60$ & $11.40$ & \multicolumn{1}{c|}{$11.33$} & & & \\
  BLEU (2009-16) & $26.85$ & $27.37$ & $27.37$ & $27.01$ & \multicolumn{1}{c|}{$27.77$} & $27.91$ & $28.34$ & $28.54$ & $28.75$ & \multicolumn{1}{c|}{$28.73$} & & & \\
  BLEU (2017) & & & & & \multicolumn{1}{c|}{} & & & & & \multicolumn{1}{c|}{$32.35$} & & & \\
  \cline{1-11}
  \multicolumn{14}{l}{English$\rightarrow$German} \\
  \hline
  Data   & P+M$_1$ & P+M$_2$ & P+M$_3$ & P+M$_4$ & P+M$_5$ & P+M$_6$ & P+M$_7$ & \multicolumn{1}{c|}{P+M$_8$} & P'+M' & P'+M' & P'+M' & P'+M' & \multicolumn{1}{c|}{P'+M'} \\ 
  Time (hours)   & $46$ & $46$ & $46$ & $46$ & $46$ & $46$ & $46$ & \multicolumn{1}{c|}{$40$} & $25$ & $25$ & $25$ & $25$ & \multicolumn{1}{c|}{$25$} \\ 
  LR      & $1.0$ & $1.0$ & $1.0$ & $1.0$ & $1.0$ & $1.0$ & $1.0$ & \multicolumn{1}{c|}{$1.0$} & $0.7^1$ & $0.7^2$ & $0.7^3$ & $0.7^4$ & \multicolumn{1}{c|}{$0.7^5$} \\
  PPL (2008)    & $12.87$ & $12.91$ & $12.38$ & $12.23$ & $12.19$ & $12.00$ & $12.26$ & \multicolumn{1}{c|}{$11.65$} & $11.51$ & $11.19$ & $10.80$ & $10.70$ & \multicolumn{1}{c|}{$10.58$} \\ 
  BLEU (2009-16)   & $21.81$ & $22.26$ & $22.52$ & $22.65$ & $22.59$ & $22.75$ & $22.79$ & \multicolumn{1}{c|}{$22.93$} & $23.35$ & $23.56$ & $23.79$ & $23.96$ & \multicolumn{1}{c|}{$24.07$} \\
  BLEU (2017) & & & & & & & & \multicolumn{1}{c|}{} & & & & & \multicolumn{1}{c|}{$26.41$} \\
  \hline
\end{tabular}
}
\end{center}
\caption{\label{tab:results2} {\it Training on parallel and synthetic data.}}
\end{table*}

\begin{figure}[h!]
   \includegraphics[width=0.48\textwidth]{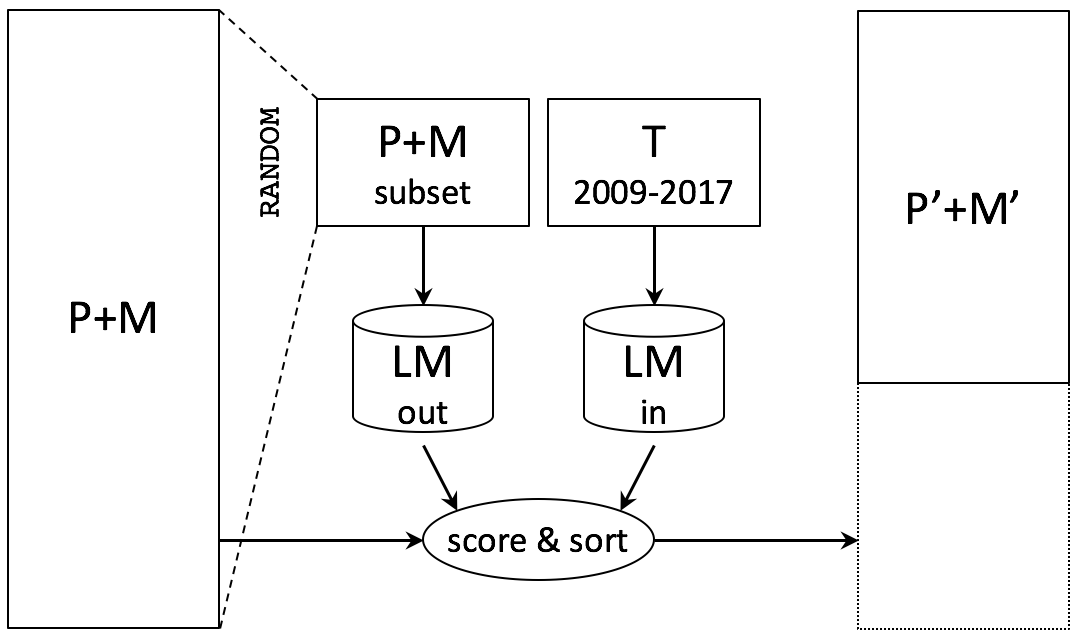}
\caption{Data selection process.}
\label{moorelewis}
\end{figure}

As for the experiments detailed in Table~\ref{tab:results1}, once all splits of the synthetic corpus were used to train our models with learning rate always set to $1.0$ ($5$ epochs for German$\rightarrow$English and $8$ epochs for English$\rightarrow$German), we began a decay mode. In this case, we decided to reduce the amount of training examples from $9$ to $5$ millions due to time restrictions. To select the training data we employed the algorithm detailed in \cite{moore10:iso}. It aims at identifying sentences in a generic corpus that are closer to domain-specific data. Figure ~\ref{moorelewis} outlines the algorithm. In our experiments, parallel and monolingual back-translated corpus are considered as the generic corpora ({\bf P+M}) while all available newstest test sets, from 2009 to 2017, are considered as the domain-specific data ({\bf T}). Hence, we aim at selecting from {\bf P+M} the closest $5$ million sentences to the newstest2009-17 data ($2.5$ from the {\bf P} and $2.5$ from the {\bf M} subsets). 

Obviously, we base our selection procedure on the source-side text of each translation direction as references for newstest2017 are not available. 

Sentences $s$ of the generic corpus are scored in terms of cross-entropy computed from two language models: a $3$-gram LM trained on the domain-specific data $H_{in}(s)$ and a $3$-gram LM trained on a random sample taken from itself $H_{out}(s)$.  Finally, sentences of the generic corpus are sorted regarding the computation of the difference between domain-specific and generic scores $H_{in}(s)-H_{out}(s)$ (score \& sort). 

\subsubsection{Hyper-specialisation on news test sets}

Similar to domain adaptation, we explore a post-process approach, which hyper-specialises a neural network to a specific domain by running additional training epochs over newly available in-domain data \cite{DBLP:journals/corr/ServanCS16}. 
In our context, we utilise all newstest sets ({\bf T}) (around $25,000$ sentences), as in-domain data and run a single learning iteration in order to fine tune the resulting network. Translations are not available for newstest2017, instead we use the single best hypotheses produced by the best performing system in Table \ref{tab:results2}. In a similar task, \cite{DBLP:journals/corr/CregoS16} report translation accuracy gains by employing a neural system trained over a synthetic corpus built from source reference sentences and target translation hypotheses. The authors claim that text simplification is achieved when translating with an automatic engine compared to reference (human) translations, leading to higher accuracy results.

Table \ref{tab:results3} details the hyper-specialisation training work. Note that the entire hyper-specialisation process was performed on approximately $6$ minutes. We used a learning rate set to $0.7$. Further experiments need to be conducted for a better understanding of the learning rate role in hyper-specialisation work. 

\begin{table}[h!]
\begin{center}
\scalebox{0.8}{
\begin{tabular}{|lcc|}
  \hline
  Epoch & 1 & 1 \\
  \hline
  \multicolumn{3}{l}{German$\rightarrow$English} \\
  \hline
  Data   & T & T-2017\\
  Time (seconds)   & $365$ & $305$ \\
  LR      & $0.7^1$ & $0.7^1$ \\
  BLEU (2017) & $32,87$ & $32,66$ \\
  \hline
  \multicolumn{3}{l}{English$\rightarrow$German} \\
  \hline
  Data   & T & T-2017\\
  Time (seconds)   & $372$ & $308$ \\
  LR      & $0.7^1$ & $0.7^1$  \\
  BLEU (2017) & $26,98$ & $26,80$ \\
  \hline
\end{tabular}
}
\end{center}
\caption{\label{tab:results3} {\it Hyper-specialisation on news test sets.}}
\end{table}

Accuracy gains are obtained despite using automatic (noisy) translation hypotheses to hyper-specialise: $+0.52$ (German$\rightarrow$English) and  $+0.57$ (English$\rightarrow$German). In order to measure the impact of using newstest2017 as training data (sefl-training) we repeated the hyper-specialisation experiment using as training data newstest sets from 2009 to 2016. This is, excluding newstest2017 ({\bf T-2017}). Slightly lower accuracy results were obtained by this second configuration (last column in Table \ref{tab:results3}) but still outperforming the systems without hyper-specialisation: $+0.31$ (German$\rightarrow$English) and  $+0.39$ (English$\rightarrow$German).

\section{Conclusions}
\label{sec:conclusions}
We described SYSTRAN's submissions to the WMT 2017 shared news translation task for English-German.
Our systems are built using OpenNMT.  We experimented using monolingual data automatically back-translated. Our resulting models were successfully hyper-specialised with an adaptation technique that finely tunes models according to the evaluation test sentences. Note that all our submitted systems are single networks. No ensemble experiments were carried out, what typically results in higher accuracy results.

\section*{Acknowledgements}
We would like to thank the anonymous reviewers for their careful reading of the paper and their many insightful comments and suggestions.

\bibliography{emnlp2017}
\bibliographystyle{emnlp_natbib}

\end{document}